# Effects of Initial Stance of Quadruped Trotting on Walking Stability

**Dongqing He** [1,2] **& Peisun Ma** [1]

[1] Research Institute of Robotics, Shanghai Jiaotong University, China
[2] School of Mechatronical Engineering, University of Petroleum, China
hedongqing2001@sohu.com

*Abstract: It is very important for quadruped walking machine to keep its stability in high speed walking. It has been indicated that moment around the supporting diagonal line of quadruped in trotting gait largely influences walking stability. In this paper, moment around the supporting diagonal line of quadruped in trotting gait is modeled and its effects on body attitude are analyzed. The degree of influence varies with different initial stances of quadruped and we get the optimal initial stance of quadruped in trotting gait with maximal walking stability. Simulation results are presented.*
*Keywords: quadruped, trotting, attitude, walking stability.*

## 1. Introduction

Quadruped machine walking in high speed requires dynamic walking with lower duty factor. Trotting gait is one of typical dynamic walking patterns. In trotting gait, two pairs of diagonal legs make standing phase respectively. Thus, there exists a moment when the body is falling down around the supporting axis, which makes the body attitude vary. Even more unfortunately, it also makes the swinging legs touch ground non-simultaneously, which will badly affect the completion of gait planning and cause gait errors. Quadruped walking machine will overbalance and fall down if gait errors accumulate to certain extent.

Inagaki and Kobayashi (Inagaki, K. & Kobayashi, H., 1994) studied the problem carefully and investigated a method of using a weight oscillator built in the center of the body to compensate the moment in sideways direction. The method is applicable if the distance from the center of gravity to the supporting axis is short. If the distance is a bit longer, the effect of compensating is unfavorable (Kurazume, R. et al., 2002). It must also be pointed out that the method does not compensate the moment in lengthwise direction and seems to be incomprehensive. Hawker and Buehler (Hawker, G. & Buehler, M. 2000) studied trotting gaits of Scout quadruped robot with a locking, unactuated knee and discussed trotting algorithms. They neglected ground contact in the simulation and used the planarizer to limit the robot's motion to the saggittal plane in the experiments. They thought it be the subject of future research to eliminate body roll about the diagonal axis. Thus, the moment falling down around the supporting axis is modeled considering both sideways direction and lengthwise direction, and the angle of rotation around the supporting diagonal line is derived in this paper. Next, the effects of the moment on body attitude and walking stability are analyzed. The analysis indicates that the degree of influence varies with different initial stances of quadruped. Finally we get the optimal initial stance of quadruped in trotting gait with maximal walking stability and simulation results are presented.

## 2. Mechanical Modelling

As the projection of gravity $G$ in the supporting plane is not on the supporting diagonal line, which causes a moment falling down around the supporting axis as illustrated in Fig. 1($\theta$ means the angel of rotation). Quadruped walking machine will rotate around the axis because of action of the moment. We want to learn how much it influences rotation and walking stability.
We assume the followings:
1. The two supporting legs on the supporting diagonal line always contact supporting ground in the 1st half of cycle period of walking and the robot only rotates around the axis without other rotations. In the 1st half of cycle period of walking, quadruped robot starts to walk and there is no obvious collision with the ground. In other periods of walking, the swinging legs will touch ground with collision and the assumption is not true probably.

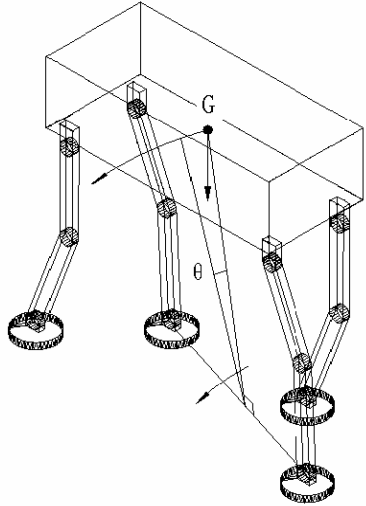

Fig. 1. Quadruped walking machine rotates around the supporting diagonal line

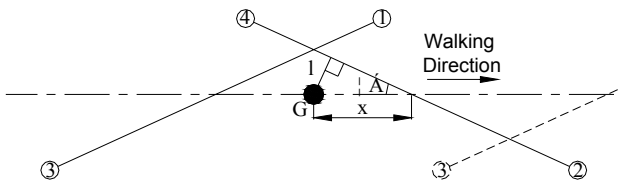

Fig. 2. Projection of the trot gait in the terrain

2. Four legs of the robot are massless and the mass of the robot is concentrated on the body.
3. The action of joint torque on the body is neglected. It is significantly smaller compared to that of the rotating moment we are analyzing because gear ratio makes them not in the same order. This assumption is very important for the next analysis.
4. Quadruped machine walks with constant speed.
5. The terrain is rigid, regular and even.
In order to illuminate dynamics relations, according to previous assumptions, especially assumption 1, 2 and 3, we have this equation by Euler's theorem in the 1st cycle period of walking:

$$mgl = I\varepsilon \quad (1)$$

where $m$ is the mass of the quadruped walking machine, $g$ is acceleration of gravity, $l$ is the distance between the projection of center of gravity(COG) in the terrain and the supporting diagonal line, $I$ is the moment of inertia of the robot and $\varepsilon$ is angular acceleration of the rotation.
It must be clarified again that (1) may be not true without assumption 1 to 3.
We obtain angular acceleration of the rotation:

$$\varepsilon = \frac{mgl}{I} \quad (2)$$

As illustrated in Fig. 2, the distance in the walking direction from the projection of center of gravity in the terrain to the supporting diagonal line is $x$. To learn about effects of stride length $S$ on walking stability, variable $x_1$ ($0 \leq x_1 \leq 0.5$) is selected and $x$ is expressed as $x_1 S$.

The constant walking speed is $v$ and $l$ is a function of time of walking $t$ because it decreases when center of gravity moves along the walking direction. According to Fig. 2, we have:

$$l(t) = \left(x_1 S - \int v dt\right) \sin \alpha \quad (3)$$

where $S$ is stride length of the robot, $\alpha$ is angle between walking direction and the supporting diagonal line.

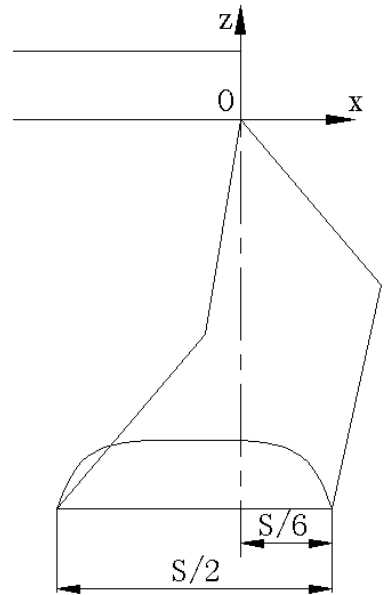

Fig. 3. Initial stance with $x = S/6$

Substituting (3) into (2), we obtain:

$$\varepsilon = \frac{mg}{I} \sin \alpha \left(x_1 S - \int v dt\right) = \frac{mg}{I} \sin \alpha \left(x_1 S - vt\right) \quad (4)$$

Integrating (4) with $t$, we obtain the angular velocity:

$$\omega = \int \varepsilon dt = \int \frac{mg}{I} \sin \alpha \left(x_1 S - \int v dt\right) dt \\ = \frac{mg}{I} \sin \alpha \left(x_1 S t - \frac{vt^2}{2}\right) \quad (5)$$

Let coefficient $A$ represent constants in equation (5), we have:

$$A = \frac{mg \sin \alpha}{I} \quad (6)$$

$$\omega = A\left(x_1 St - \frac{vt^2}{2}\right) \quad (7)$$

Integrating (7) with $t$, we obtain the angle of the rotation:

$$\theta = A\left(\frac{x_1 St^2}{2} - \frac{vt^3}{6}\right) \quad (8)$$

Substituting $v = S/T$ into (8), we have:

$$\theta = AS\left(\frac{x_1 t^2}{2} - \frac{t^3}{6T}\right) \quad (9)$$

When the 1st half of cycle period of walking ends and then the next pair of legs start to be in standing phase ($t = T/2$), the angle of rotation around the supporting diagonal line is:

$$\theta = AST^2\left(\frac{x_1}{8} - \frac{1}{48}\right) \quad (10)$$

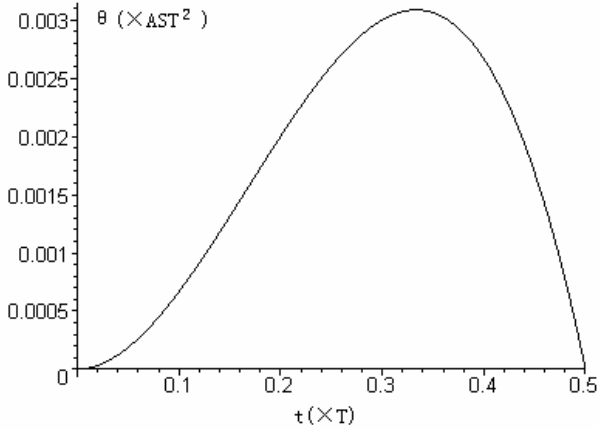

Fig. 4. $\theta$ $t$ curve with $x = S/6$

From (10), we can see that variable $x_1$ affects the angle of rotation around the supporting diagonal line. It is easy to see that the angle of rotation $\theta$ may be 0 if $x_1$ is equal to $1/6$. Next we will analyze how does variable $x_1$ influence the angle of rotation and walking stability largely.

## 3. Analysis

From (10), we can see that:
1. The angel of rotation $\theta$ is proportional to $ST^2$. The conclusion is also coincident with the experiment data made by Kimura et al. (Kimura, H. 1988) with collie-2. From $S = vT$ we can see that $\theta$ is proportional to $T^3$ with constant walking speed. In order to maintain attitude stability and walking stability, the smaller the cycle period and stride length, the better.

2. The angel of rotation $\theta$ is also proportional to coefficient $A$. From (6), $A$ is proportional to $m\sin\alpha$ and inverse proportional to $I$. So a minor $m$ and a rather large $I$ are advantaged for decreasing the angel of rotation $\theta$ and reinforcing stability.

3. The initial position $x$ ($x_1 S$) of quadruped walking machine largely influences attitude stability and walking stability of the robot. If $x$ is $S/6$ ($x_1 = 1/6$, see Fig. 3), the angle of rotation at the end of supporting will be 0 theoretically (see Fig. 4) and the body of the robot is upright without any inclination. It is very advantaged for changing of supporting legs and dynamic stable walking. If $x$ is another value, such as 0, $S/4$ and so on, the angle of rotation around the diagonal supporting diagonal line is not 0, it will aggravate the collision with the ground and is not advantaged for dynamic stable walking.

## 4. Simulation Results

To verify the analysis, ADAMS virtual prototype software is used to model quadruped walking machine (see Fig. 5). To implement a simple straight-line walk, there are 3 joints around pitch axis for each leg of the robot. The ankle joint is a passive joint in order to increase walking stability. The cycle period $T$ of the walk is 0.7 s. To eliminate the effects of collision with the ground on simulation results, walk starts at 0.7 s in the simulation. In order to study the effects of various initial stances on walking stability, different values of initial positions $x$ are selected, such as 0, $S/6$ and $S/4$. Other parameters are the same for the model in simulations.

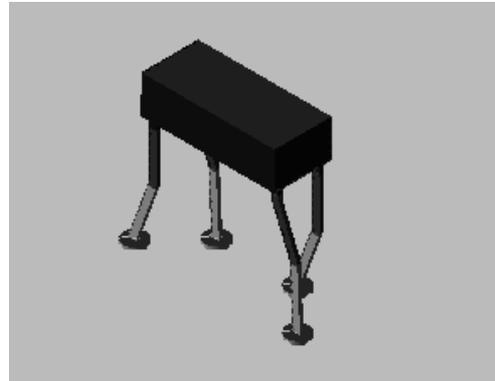

Fig. 5. ADAMS model of quadruped walking machine

By a series of simulations we get data of roll angle and pitch angle at $t = T/2$ in walking, the number of stable walking cycle periods and the status of variation of those angles in walking (see Table 1 & Fig. 6).

From simulation results, at the end of the 1st half of cycle period of walking, variation of body attitude is the smallest if initial position $x$ is $S/6$ (see Table 1. Here, it refers to the data of roll angle and pitch angle). Besides, the number of stable walking periods is the maximum in all initial positions (see Table 1 & Fig. 6). The results illustrate that initial stance do influence walking stability of quadruped walking and $x = S/6$ is the optimal initial stance. Here, the optimal initial stance refers to the stance with minimal attitude varying at the end of the supporting ($t = T/2$) and maximal walking stability.

| Initial position | Roll angle ($t = T/2$) | Pitch angle ($t = T/2$) | Number of stable periods |
|---|---|---|---|
| 0 | 2.3° | 1.7° | 8 |
| $S/6$ | 0.5° | 3.1° | >19 |
| $S/4$ | 0.7° | 5.0° | 15 |

Table 1 Simulation results of different initial stances of quadruped trotting

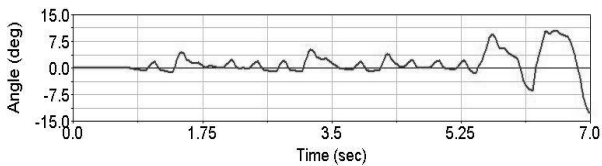

(a) Roll angle and pitch angle with $x = 0$ (the upper is roll angle and the lower is pitch angle)

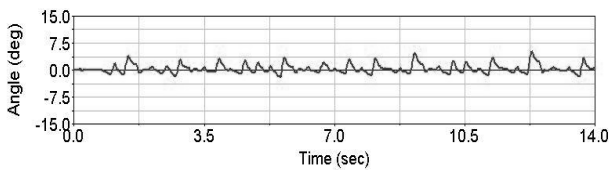

(b) Roll angle and pitch angle with $x = S/6$

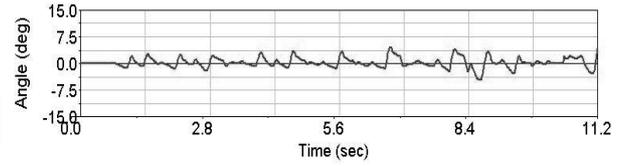

(c) Roll angle and pitch angle with $x = S/4$

Fig. 6. Roll angle and pitch angle of the robot in walking (from the beginning of 0.7 s)

## 5. Conclusion

In this paper we presented that initial stance of quadruped robot in trotting gait largely influences walking stability. Through equations and simulations we found that the angle of rotation around the supporting diagonal line of quadruped in trotting gait is minimal and walking stability is maximal if initial position $x$ is $S/6$.

The optimal initial stance also makes it easier to adjust and control body attitude of the robot. If buffer gear and vibration absorber are applied and proper control is utilized, quadruped walking machine will walk more quickly and stably.

It must also be pointed out that the value of optimal initial stance in this paper is true with constant walking speed. If there is any acceleration in walking, the optimal initial stance will change. But the method used in this paper could be used to get a new optimal initial stance.